\documentclass[10pt,twocolumn,letterpaper]{article}

\usepackage{authblk}
\usepackage{wacv}
\usepackage{times}
\usepackage{epsfig}
\usepackage{graphicx}
\usepackage{array}
\usepackage{amsmath}
\usepackage{amssymb}
\usepackage{bbm}
\usepackage{color,soul}
\usepackage{algorithm}
\usepackage{algpseudocode}
\usepackage[inline]{enumitem}
\usepackage{multirow}
\usepackage{wrapfig}
\usepackage{subcaption}
\usepackage{stmaryrd}
\usepackage[font=small,labelfont=bf,tableposition=top]{caption}
\newcommand{\ra}[1]{\renewcommand{\arraystretch}{#1}}
\usepackage{float}

\DeclareMathOperator{\One}{\mathbbm{1}}
\usepackage{authblk}
\usepackage[utf8]{inputenc} 
\usepackage[T1]{fontenc}    
\usepackage{hyperref}       
\usepackage{url}            
\usepackage{booktabs}       
\usepackage{amsfonts}       
\usepackage{nicefrac}       
\usepackage{microtype}      

\wacvfinalcopy 

\def\wacvPaperID{742} 

\ifwacvfinal\fi
\begin{document}

\title{Calibrated Domain-Invariant Learning for Highly Generalizable Large Scale Re-Identification}


\newcommand*\samethanks[1][\value{footnote}]{\footnotemark[#1]}
\author[1]{Ye Yuan\vspace{-0.8em}}
\author[1]{Wuyang Chen}
\author[1]{Tianlong Chen}
\author[2]{Yang Yang}
\author[3]{Zhou Ren}
\author[1]{Zhangyang Wang}
\author[3]{Gang Hua}
\affil[ ]{\textit {\{ye.yuan, wuyang.chen, wiwjp619, atlaswang\}@tamu.edu,}} 
\affil[ ]{\textit {yang.yang2@walmart.com, \{renzhou200622, ganghua\}@gmail.com}}
\affil[1]{Department of Computer Science and Engineering, Texas A\&M University}

{\makeatletter
\renewcommand\AB@affilsepx{, \protect\Affilfont}
\makeatother
\affil[2]{Walmart Technology} 
\affil[3]{Wormpex AI Research}}

\affil[ ]{\small \url{https://github.com/TAMU-VITA/ADIN}}
\renewcommand\Authands{ and }

\maketitle
\ifwacvfinal\thispagestyle{empty}\fi

\begin{abstract}
    Many real-world applications, such as city scale traffic monitoring and control, requires large scale re-identification. However, previous ReID methods often failed to address two limitations in existing ReID benchmarks, {\em i.e.},  {\em low spatiotemporal coverage} and {\em sample imbalance}. Notwithstanding their demonstrated success in every single benchmark, they have difficulties in generalizing to unseen environments.  As a result, these methods are less applicable in a large scale setting due to poor generalization.  
    In seek for a highly generalizable large-scale ReID method,
    we present an adversarial domain-invariant feature learning framework (\textbf{ADIN}) that explicitly learns to separate identity-related features from challenging variations, where for the first time ``free'' annotations in ReID data such as video timestamp and camera index are utilized.
    Furthermore, we find that the imbalance of nuisance classes jeopardizes the adversarial training, and for mitigation we propose a calibrated adversarial loss that is attentive to nuisance distribution. Experiments on existing large-scale person/vehicle ReID datasets demonstrate that ADIN learns more robust and generalizable representations, as evidenced by its outstanding \textbf{direct transfer} performance across datasets, which is a criterion
    that can better measure the generalizability of large scale Re-ID methods.
\end{abstract}

\section{Introduction}
The increasing usage of sensors, especially surveillance cameras, in smart communities and cities, has resulted in immense opportunities in developing large-scale computer vision and machine learning algorithms for Urban Informatics. As a core algorithmic component of many camera-based applications, the re-identification of subjects across multiple cameras, known as ReID, has been one of the most demanded capabilities. For example, person ReID targets to match and return images of a probe person from a large-scale gallery set collected from different cameras, which is an important in security and surveillance. As another example, vehicle ReID uses traffic cameras as citywide sensors to optimize flows and manage traffic accidents. These ReID problems have recently drawn explosive attention from both academia and industry. Despite notable research progress, there remain to be major gaps between the research efforts and the practical needs in large-scale deployment.

\subsection{The Generalizability Gaps: Low Coverage and Sample Imbalance} \label{sec:intro}

\begin{figure*}[!h]
\begin{center}
    \includegraphics[width=.355\linewidth]{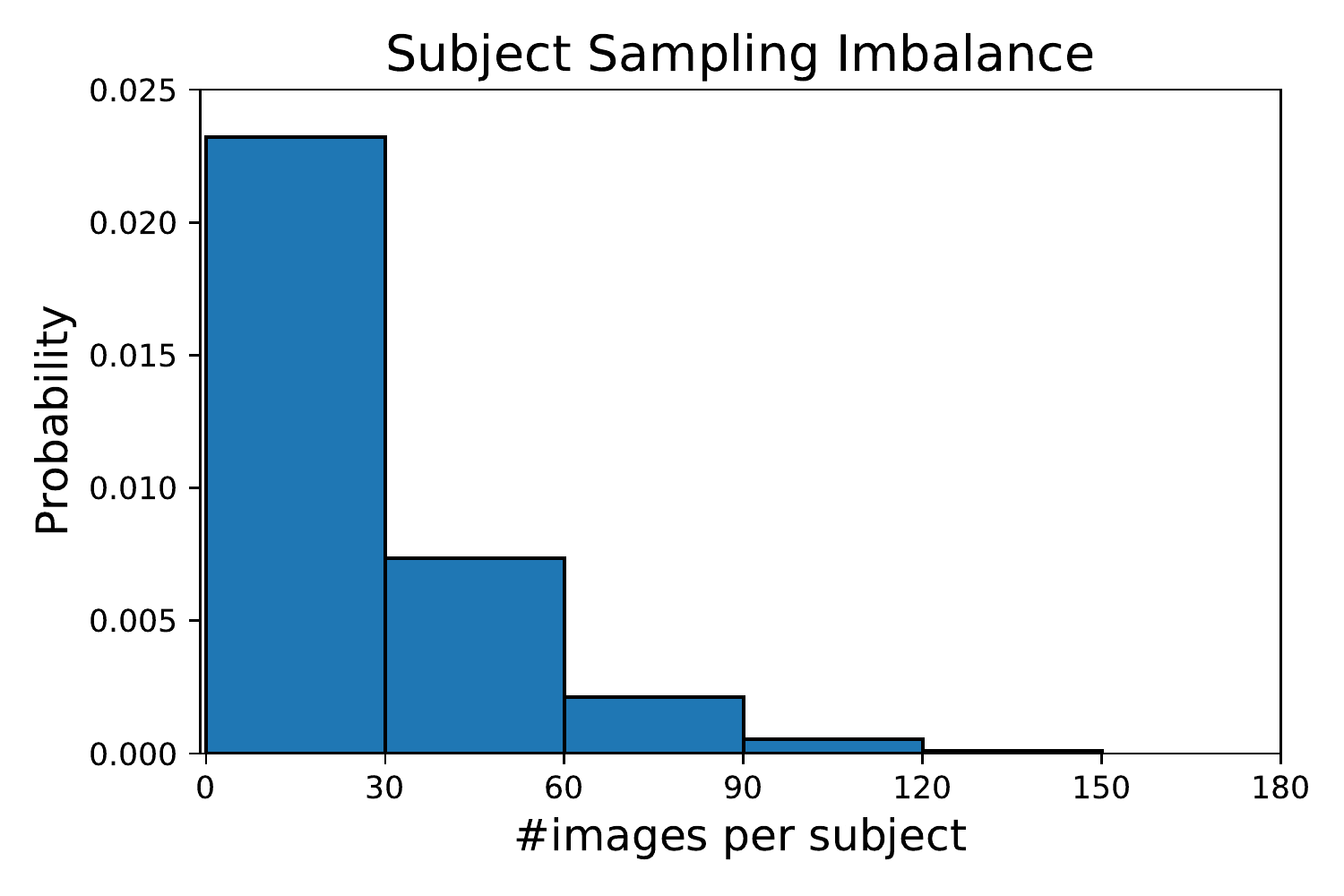}
    \includegraphics[width=.355\linewidth]{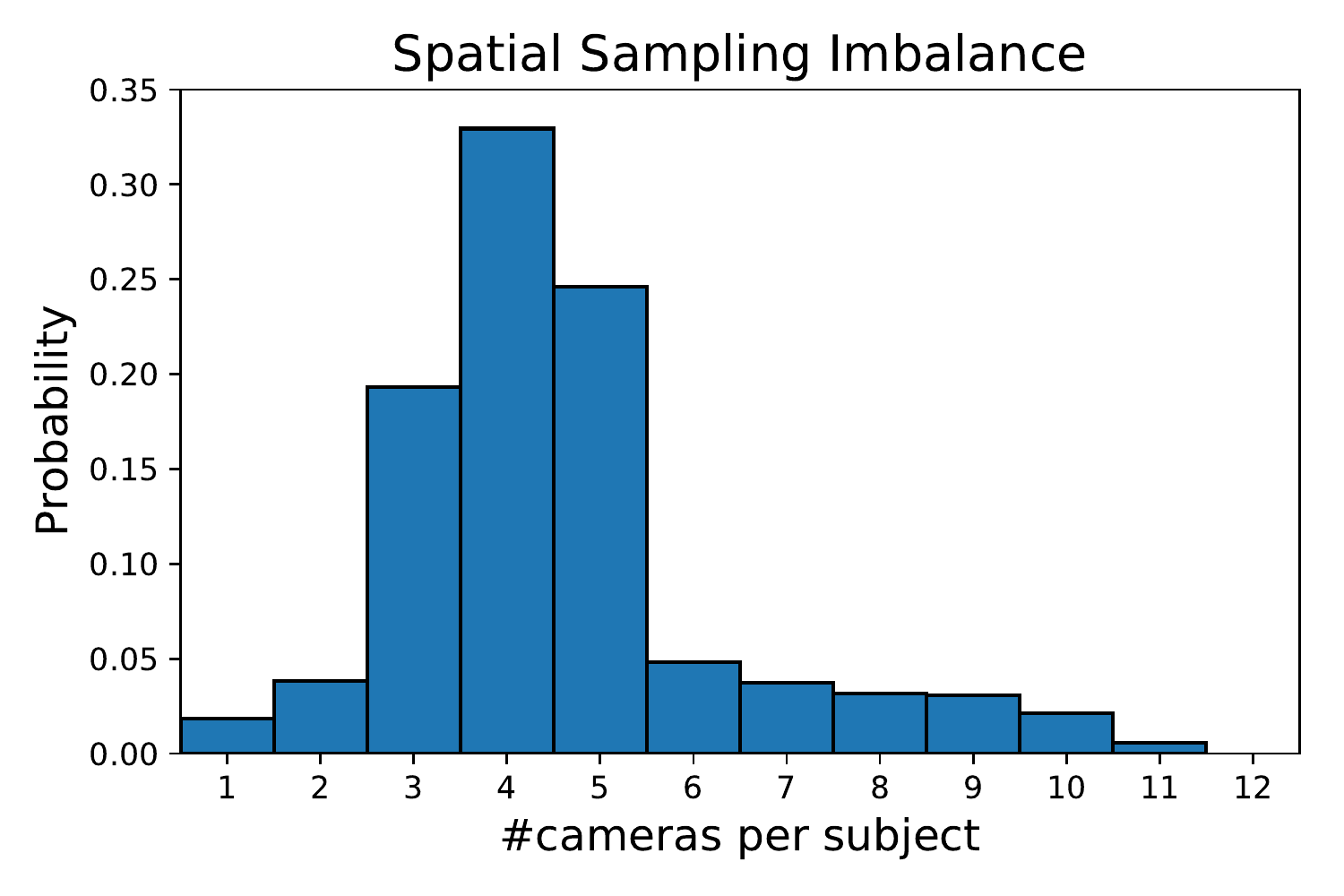}
    \includegraphics[width=.24\linewidth]{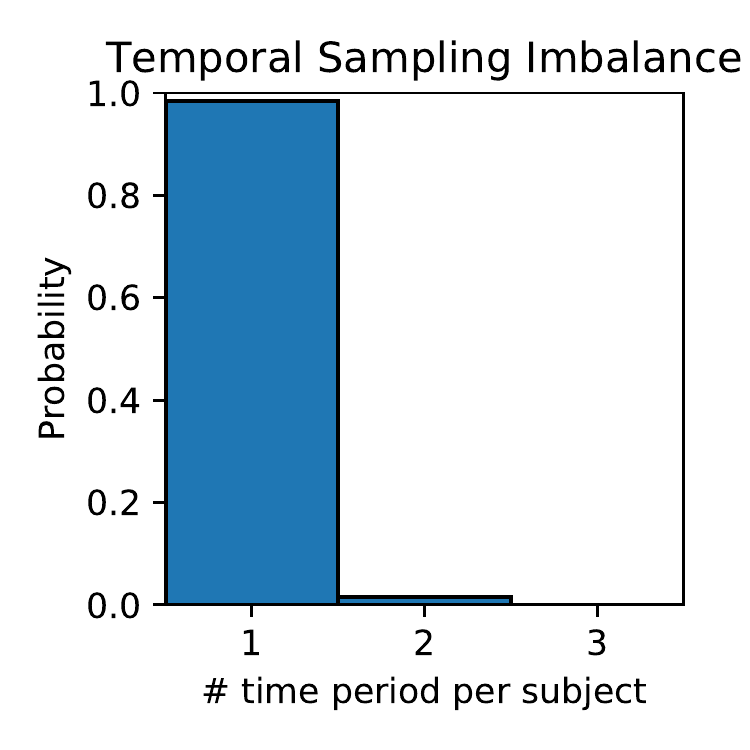}
\end{center}
\vspace{-1em}
\caption{The sample imbalance issue in ReID datasets (MSMT17 \cite{Wei_2018_CVPR} as an example). The imbalance manifests in: (left) histogram of image numbers per subject; (middle) histogram of camera numbers in which each subject was captured; (right) histogram of time period numbers (1/2/3: morning/noon/afternoon) during which each subject was captured. The latter two histograms show to be clearly skewed towards the lower end, implying the ``localized'' patterns of person activities in MSMT17. Similar observations can be found in other peer datasets.}
\vspace{-1em}
\label{fig:sample_imbalance}
\end{figure*}

\textbf{High generalizability is essential to deploy ReID at large scale.} ReID is by nature an ``open-world'' problem, which is necessitated to generalize in two aspects. First, the ReID models have to generalize to subjects (\textit{e.g.}, person or vehicle) that were unseen in the training set. Second, ReID shall also generalize to re-identifying subjects that undergo unseen variations induced by changes in background, illumination, viewpoint -- \textit{i.e.}, generalizing to unseen ``combinations'' of subjects and accompanied variations. We call those unwanted variations as \textbf{nuisances} since they are irrelevant to the subject identities.

With the rapidly expanding smart city/community camera networks, good generalizability becomes the \textit{ultimate goal} for the practical deployment of large-scale ReID systems: it has to stay effective to new subjects, can scale up to new locations, and be reliable over time. However, most existing ReID algorithms may not have addressed these well, which may hinder their deployment in practice.

\textbf{Gap \#1: Low Spatiotemporal Coverage.} Existing ReID datasets are limited in data volume. They are restricted in not only subject numbers but also spatial/temporal coverage, making them oversimplified in reference to the complexity and diversity of the large-scale scenes. A recent study \cite{camera2014} showed that in 2014, there were 125 video surveillance cameras per thousand people in the U.S.; whereas most ReID datasets were collected only from 10 or fewer cameras (see Table \ref{table:dataset}). What is worse, most datasets hand-picked video frames captured in similar outdoor environments and/or under normal lighting conditions; however, practical ReIDs need to cope with drastically diverse locations, indoor-outdoor matching, intensive day-long illumination variations, and more.

\textbf{Gap \#2: Sample Imbalance}\footnote{In ReID, ``sample imbalance'' traditionally refers to the negative samples being much more than positive ones during pairwise training \cite{wang2018person}. We are using this term in an apparently different context.}. We believe this issue remains yet \textit{overlooked} in the ReID community, especially when the open scenarios in real cities go beyond the controlled conditions in training datasets. First, a ReID dataset consists of images from different subjects, where every subject has an indefinite number of images (Fig.\ref{fig:sample_imbalance} left), making the subject class distribution \textit{non i.i.d}. Similarly, different nuisances may also appear in a dataset with different frequencies. \underline{What is worse}, a subject may (very likely) be only captured by a small portion of cameras (Fig.\ref{fig:sample_imbalance} middle) within a small portion of time periods (Fig.\ref{fig:sample_imbalance} right). Intuitively, a person or a vehicle usually appears most in certain regions within certain hours, rather than being a wanderlust anywhere anytime ``uniformly'' in a city. Hence, the conditional distribution of nuisances given a subject is also \textit{heavily non i.i.d}: some subjects may display strong yet superficial correlations with some nuisances, making it highly challenging to decouple them. 

Considering these gaps, we argue that evaluating (and even overfitting) on single datasets is not helpful for designing algorithms that would be practically deployed in real-life large-scale scenarios.

\subsection{Our Solution and Contributions}
\begin{figure*}[!h]
\begin{center}
    \includegraphics[width=.65\linewidth]{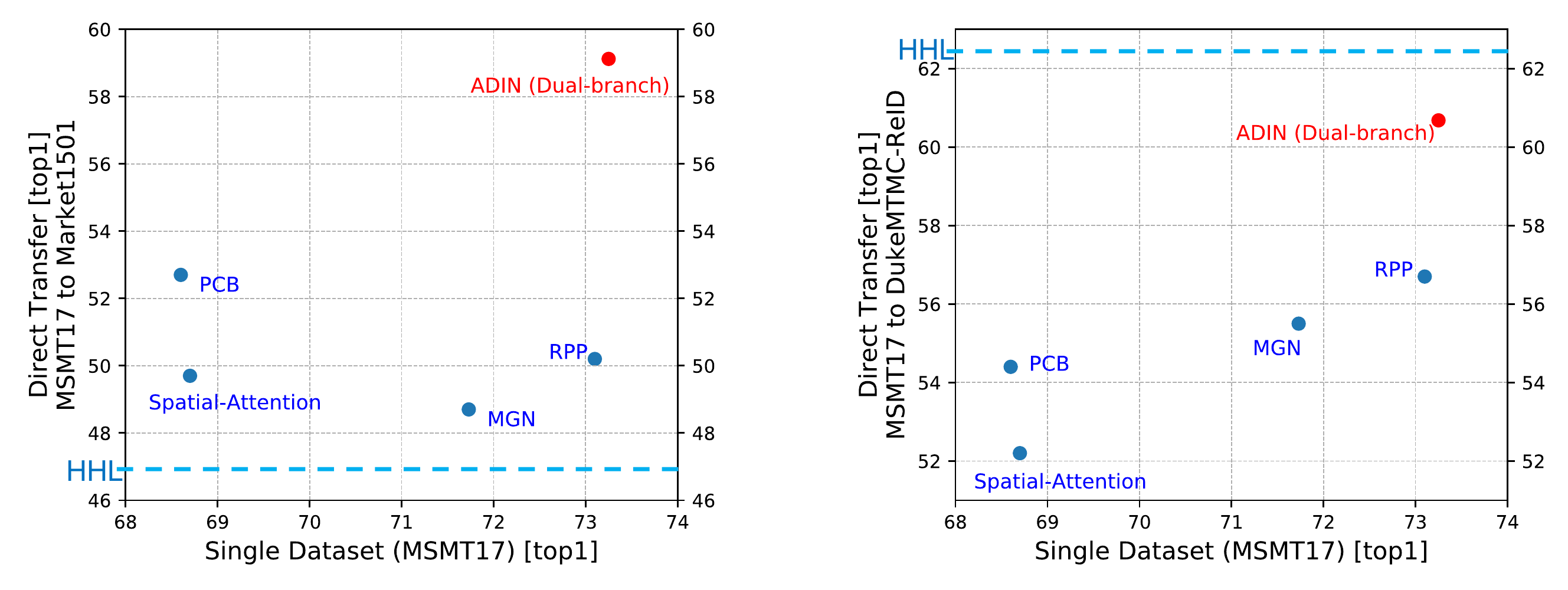}
\end{center}
\vspace{-1em}
\caption[foo bar]{Top1 accuracy on a single-dataset (MSMT17 \cite{Wei_2018_CVPR}) and direct transfer from MSMT17 to Market1501 \cite{Zheng_2015_ICCV} (left) and to DukeMTMC-ReID \cite{Ristani_2016_ECCV, Zheng_2017_ICCV} (right). In contrast to our ADIN (red dot in top-right) which achieved competitive performance on both a single dataset and direct transfer, we find other (single-dataset) top-performers suffer from very poor generalizability to unseen domains, indicating the misaligned goal between overfitting small-scale single dataset and generalizing to large-scale unseen scenarios in real life. See section \ref{sec:direct_transfer} for details. Methods studied: Spatial-Attention \cite{wang2018parameter}, PCB\cite{Sun_2018_ECCV}, RPP \cite{Sun_2018_ECCV}, MGN \cite{wang2018learning}, HHL\footnotemark \cite{Zhong_2018_ECCV}.}
\vspace{-1em}
\label{fig:scatter_plot}
\end{figure*}

The low spatiotemporal coverage coupled with the inherent sample imbalance, have placed jeopardy for practical large-scale ReID. Existing ReID algorithms trained on a single dataset are prone to overfitting nuisances of the training set, and therefore suffering from poor generalizability, as indicated by poor direct transfer performance to unseen datasets (Fig. \ref{fig:scatter_plot} and Table \ref{table:direct_transfer_msmt2duke.market}).

Fortunately, most scene-related nuisances are caused by \textit{camera-specific} and/or \textit{time-specific} factors (Fig. \ref{fig:sample_imbalance} middle and right). Since \textit{video timestamp} or \textit{camera index} are freely available in video surveillance as metadata and are provided by almost all existing person/vehicle ReID datasets, they can be potentially utilized as auxiliary supervision, although few image-based ReID methods have taken advantage of them.

This paper aims to improve the \textbf{generalizability} of ReID models in large-scale settings, by resorting to a novel domain-invariant feature learning perspective. Inspired by \cite{wu2018towards, wang2019privacy}, we consider samples (of different subjects) with the same nuisance to be from one domain (such as images captured by the same fixed camera, or in the same time period). This is because scene-related changes (background, illumination, viewpoint, etc.) heavily dominate the appearances of images. Different types of nuisances hence becomes domain-specific features. In contrast, one subject can be captured at different cameras and time periods, and the subject's identity features should apparently remain domain-invariant. Therefore, our main idea is to extract ReID features that can: (1) be utilized to faithfully classify subjects into correct classes; (2) be resilient and invariant to those identified nuisances -- in other words: no reliable classifier can be trained on those features to predict those nuisances.

\footnotetext{HHL uses images from both source and target domain for domain adaptation, and thus has no single-dataset performance. We use a horizontal line to represent its domain adaptation performance.}

In addition to setting up a new goal of generalizability, our contributions include a corresponding new algorithm as well as a new evaluation criterion. We formulate our adversarial domain-invariant learning framework (\textbf{ADIN}), by taking advantage of ``free'' annotations like video timestamp and camera index, to separate identity-related features from scene-specific nuisances. To our best knowledge, we are the first to utilize those ``free'' annotations for image-based ReID, to effectively suppress the overfitting of nuisances. Moreover, we find the imbalance of nuisance distribution w.r.t. subjects hampers the adversarial learning. A novel calibrated adversarial loss is therefore introduced to tackle the nuisance class imbalance for ADIN. Measured by a new \textbf{direct transfer} performance criterion on popular large-scale ReID benchmarks, our ADIN demonstrates outstanding generalizability and outperform previously reported results and even some that rely on domain adaptation using target data (Fig.\ref{fig:scatter_plot}).


\section{Related Work} \label{sec:related_work}
\subsection{ReID Datasets} \label{sec:datasets}
\begin{table}
\centering \arraybackslash
\caption{Publicly available benchmarks for person/vehicle ReID: the number of cameras, identities, and average bounding boxes per identity.}
{\scriptsize
\ra{1.0}
\begin{tabular}{ccccc}
\toprule
                         & Benchmark      & \#cameras         & \#ID  & \#boxes/ID            \\ \midrule
\multirow{4}{*}{\rotatebox[origin=c]{90}{person}}  & Market-1501 \cite{Zheng_2015_ICCV}   & 6                  & 1501  & 21.5                     \\ 
                         & DukeMTMC-ReID\cite{Ristani_2016_ECCV, Zheng_2017_ICCV} & 8                  & 1812  & 20.1                     \\ 
                         & CUHK03\cite{li2014deepreid}        & 2                  & 1467 & 9.0                      \\ 
                         & \textbf{MSMT17}\cite{Wei_2018_CVPR}        & \textbf{15}                 & \textbf{4101}  & \textbf{30.8}                     \\ \midrule 
\multirow{3}{*}{\rotatebox[origin=c]{90}{vehicle}} & VeRi\cite{liu2016large}          & 20                 & 619 & 64.6                       \\ 
                         & VehicleID\cite{liu2016deep2}     & 2                  & 26267  & 8.4                    \\ 
                          & \textbf{VeRi-776}\cite{liu2016deep}      & \textbf{20}                 & \textbf{776}  & \textbf{63.6}                      \\ 
\bottomrule
\end{tabular}
}
\label{table:dataset}
\end{table}

The disconnection between research-level datasets and community/city-level video warehouse remains to hinder the real-life applications of ReID. Table \ref{table:dataset} summarizes mainstream person/vehicle ReID datasets. For person ReID, considering that even in a grocery store there are usually dozens even more than 100 cameras \cite{camera_amazon} and over 550 visitors per day \cite{visitor_amazon}, current datasets are more or less overly simplistic. More specifically, the Market-1501 \cite{Zheng_2015_ICCV}, DukeMTMC-ReID \cite{Ristani_2016_ECCV, Zheng_2017_ICCV} and CUHK03 \cite{li2014deepreid} are all collected in small outdoor regions, and in short time periods (usually well-lighted daytime). The latest MSMT17 dataset \cite{Wei_2018_CVPR} led positive progress towards real large-scale usage, by including geo-spatially diverse cameras (both indoor and outdoor) and varying time periods (morning, noon and afternoon) and illuminations.

Vehicle ReID witness similar situations, where exiting benchmarks' scale and diversity are still far from being comparable to reality.  Previous datasets such as VehicleID \cite{liu2016deep2} have small camera or vehicle numbers, as well as limited viewpoints. A recent VeRi-776 dataset \cite{liu2016deep} presented a relatively realistic benchmark with cameras spanning a large spatial coverage and other variations, which is one-step close to being representative for large-scale vehicle ReID.

\subsection{ReID Evaluation Metrics}
 The standard ReID pipeline picks a dataset, learning the model from its training set and evaluating the model's retrieval accuracy or mean average prevision (mAP) on the held out testing set (with non-overlapping subjects). However, this \textbf{single-dataset evaluation is often insufficient in reflecting true generalizability} (Fig.\ref{fig:scatter_plot}) since they overlook a fact, \textit{i.e.}, due to the low coverage of most datasets, the training and testing sets of the same ReID dataset tend to be highly similar in terms of spatiotemporal nuisances (even overlapping or sharing camera IDs). Therefore, a high accuracy/mAP on the same testing set may be misleading, as that could be a result of nuisance overfitting. 

In view of the above, increasing attention has been paid to domain adaption in ReID recently, \textit{i.e.}, training on one source dataset, tuning the trained model on some different target domain data, and finally evaluating model accuracy/mAP on the target dataset. Domain adaptation methods \cite{Zhong_2018_ECCV} emphasize the generalizability of ReID to new data. Unfortunately, they require target domain data (sometimes even auxiliary attribute annotations in target domain \cite{Wang_2018_CVPR}) for re-training purposes. Considering the city growth as well as the explosive increase of cameras, it is unrealistic to collect new data and re-train ReID models for every new domain (\textit{e.g.}, a new camera or a group of cameras in a local region), making it non-trivial for domain adaptation to scale up. 

In contrast, we advocate a far more challenging but practically evaluation criterion: \textbf{direct transfer} performance across datasets, to measure ReID model generalizability (see section \ref{sec:direct_transfer}).

\subsection{Algorithms for Improving ReID Generalizability}
\textbf{Data Augmentation for ReID.} Tian et al. \cite{Tian_2018_CVPR} proposed to generate images of the same identity with different random backgrounds. In Ma et al. \cite{Zhong_2018_CVPR} a camera-invariant descriptor subspace is learned and the camera styles are transferred to each sample. \cite{Ma_2018_CVPR} first learned a set of disentangled foreground, background and pose factors, then re-composed them into novel samples. Data augmentation will amplify the training burden. Moreover, they still suffer when transferred to an unseen dataset, since no single training dataset (as being low-coverage itself) can cover sufficient real-world variations.

\textbf{Domain Adaptation for ReID.} Deng et al. \cite{Deng_2018_CVPR} proposed to generate a new training set of images whose identities were from the labeled source domain, while the camera styles were translated from the unlabeled target domain. Zhong et al. \cite{Zhong_2018_ECCV} introduced a Hetero-Homogeneous Learning (HHL) method to learn person embedding with camera variances and domain connectedness, through inter- and intra-domain pairwise contrastive learning. An unsupervised image translation approach was presented in \cite{peng2019cross} for source models to learn the style of the target domain. Different from them, our ADIN is ``directly transferable'' to unseen domains, without seeing or re-training on target data.


\section{The Proposed Approach: ADIN}
\subsection{Problem Formulation}
Given a training image $X$ with the identity labels $Y_I$ and the (freely) available nuisances label $Y_N$ (one or multiple, such as camera ID, video timestamp, etc.), our goal is to learn a feature representation $f_E(X)$ that is highly \textbf{relevant} to the identity label, yet being invariant or \textbf{irrelevant} to the nuisances label. Using a function $R$ to represent the correlation between the feature and the label, our learning goal is mathematically described as: 
\begin{equation}
\begin{split}
R(f_E(X), Y_I) \approx R(X,Y_I),\text{\space\space\space} R(f_E(X), Y_N) \ll R(X,Y_N).
\end{split}
\label{object_1}
\end{equation}
We adopted an identity prediction module $f_I$ which projects the feature $f_E(X)$ into identity-related features, and a nuisance prediction module $f_N$ that extracts scene-specific nuisances from $f_E(X)$. Without loss of generality, both of them are assumed to have softmax-form outputs. Note that $f_E$, $f_I$ and $f_N$ all need to be learned together. Their interactions provide mutual supervision. In particular, $f_N$ will serve as an ``adversary'' role.

To evaluate $R$ practically, a straightforward choice is to use two standard classification-oriented loss functions $L_I$ and $L_N$ (\textit{e.g.}, cross-entropy) for $f_I$ and $f_N$ respectively and minimize the classification error rate of $Y_I$ from $f_E(X)$, while maximizing the classification error rate of $Y_N$ from $f_E(X)$. Our task then becomes to simultaneously train $f_E$, $f_I$ and $f_N$, so as to minimize $L_I(f_I(f_E(X)),Y_I)$ meanwhile maximizing $L_{N}(f_N(f_E(X)), Y_N)$. 
\begin{equation}
\begin{split}
    \min_{\substack f_E} L_I(f_I & (f_E(X)),Y_I),\text{\space\space\space}
    \max_{\substack f_E} L_{N}(f_N(f_E(X)), Y_N).
\end{split}
\label{object_2}
\end{equation}
Maximizing $L_{N}(f_N(f_E(X)), Y_N)$ is not straightforward to implement. Previous work \cite{ganin2014unsupervised} reversed the sign of gradient computed from minimizing $L_{N}(f_N(f_E(X)), Y_N)$, \textit{i.e.}, using gradient ascent. However, we observed in experiments that the reverse gradient approach yielded unstable training process. Instead, we introduce a new $L_{adv}$ loss, to encourage the \textit{disparity} between $f_N(f_E(X))$ and $Y_N$: a smaller $L_{adv}$ value is expected to indicate a \textit{worse} correlation between them. 
A detailed discussion about the choice of $L_{adv}$ will be presented in section \ref{sec:loss}.

Finally, the training goal of ADIN is represented below ($\beta >0$ is a scalar):
\begin{equation}
\begin{split}
    \min_{\substack f_E} L_I(f_I & (f_E(X)),Y_I)
    + \beta L_{adv}(f_N(f_E(X)), Y_N).
\end{split}
\vspace{-1em}
\label{object_3}
\end{equation}
Meanwhile, in order to keep adversarial domain-invariant feature learning effective so as to learn meaningful $f_E$, we need to also maintain $f_N$ to be a strong competitor. That implies a \textit{hidden constraint}, \textit{i.e.}, avoiding $L_{N}(f_N(f_E(X)), Y_N)$ growing large too quickly, in which case $f_N$ becomes to have too poor nuisance classification ability so that it cannot make a useful adversary. 

\begin{figure*}[!ht]
\begin{center}
    \includegraphics[width=.75\linewidth]{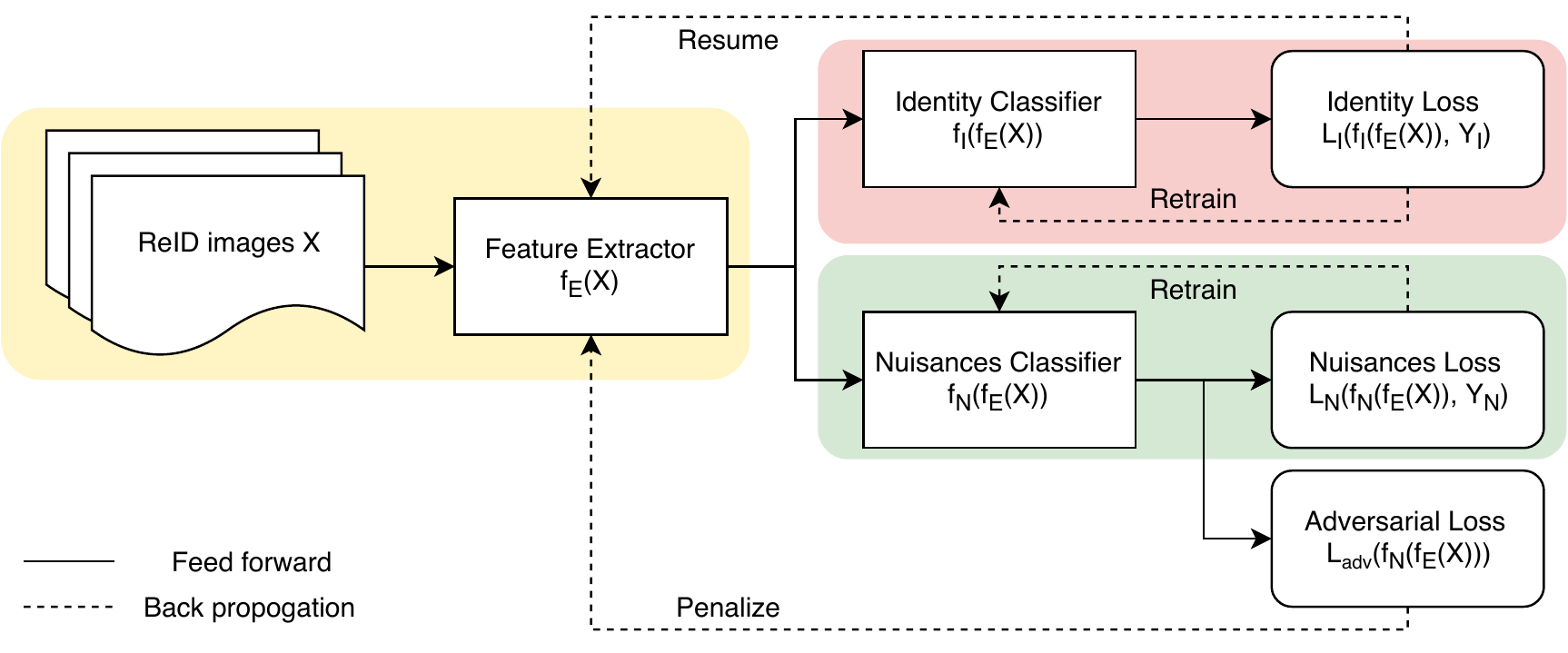}
\end{center}
\vspace{-1em}
\caption{Overview of the ADIN framework, and illustration of its training strategy. See section \ref{sec:training} for details.}
\vspace{-1em}
\label{fig:model}
\end{figure*}

\subsection{Calibrated Adversarial Loss for Imbalanced Nuisances}\label{sec:loss}
As noted in section \ref{sec:intro}, both subject and nuisance classes (conditioned on the subject) suffer from sample imbalances. We experimentally observed the subject imbalance to have less severe impact on ReID performance (\textit{e.g.}, comparing using standard and reweighted softmax loss), and therefore keep using a standard softmax function for $L_I$. However, the nuisance class imbalance was found to cause considerable training instability and performance degradation for the adversarial learning. We thus focus a detailed discussion on how we derive a robust $L_{adv}$ for the imbalanced nuisances. 

We denote $c = [c_1, ..., c_K]$ as the softmax-form output of $f_N$, where $K$ is the total nuisance class number. We next present three options that we tried for $L_{adv}$, among which our proposed new Option \#3 is experimentally validated to be the best choice for ADIN (see section \ref{sec:ablation} for details).

\noindent{\bf Option \#1: Reverse Gradient (RG)}. One possibility is to adopt the reversal gradient layer \cite{Yaroslav_2016_JMLR}. It computes the gradient for minimizing the cross-entropy between $f_N(f_E(X))$ and $Y_N$, then reversing the gradient sign.  
However, this objective becomes problematic in our case, as it was observed to cause large fluctuations in the training curve and failure of convergence. Moreover, when both $f_N$ and $f_E$ are initialized from pre-trained models (practically improving convergence and results), the gradients start with very small magnitudes and the model updates become too slow. RG is written as ($Y^*$ is the true label):
\begin{equation}
\begin{aligned}
    L_{adv} (X, Y_N) = -L_N = \sum_{\substack k=1}^{\substack K} \One_{[k=Y^*]} \log (c_k) \\
\end{aligned}   
\label{grad_reverse}
\end{equation}
\noindent{\bf Option \#2: Negative Entropy (NE). } An alternative is to minimize the negative entropy function of the softmax vector (or equivalently, its \textit{cross-entropy} with uniform distribution), as to encourage ``uncertain'' predictions of nuisance attributes (\textit{e.g.}, camera ID and video timestamps) from the extracted ReID features.  The rationale is that, if the nuisance prediction is only as good as the random guess (uniform distribution over all classes), then the feature is not informed of nuisances and therefore can generalize to unseen nuisances. NE could be written as
\begin{equation}
\begin{split}
    L_{adv} (X, Y_N) & = \sum_{\substack k=1}^{\substack K} c_k \log(c_k).
\end{split}
\label{KL_div}
\end{equation}
Importantly, although $Y_N$ does not explicitly occur in the loss form, it will still be utilized in re-training $f_N$ to make a sufficiently strong competitor (section \ref{sec:training}). We previously also tried the KL Divergence and the Jensen-Shannon Divergence between the softmax and uniform distribution, but NE appears to work best in practice. 

\noindent{\bf Option \#3 (Proposed): Calibrated Negative Entropy Loss (CaNE).} \label{loss:cane} Despite boosting uncertainty, NE overlooks the practical imbalance of nuisance class distribution w.r.t. subjects. A well-known solution is to add a modulating factor to cross-entropy loss, ensuring that the majority class/easy decisions do not overwhelm the loss \cite{lin2017focal}. We propose a reweighted form of NS, called Calibrated Negative Entropy Loss (CaNE), to make $L_{adv}$ attentive to the skewed nuisance distribution
\begin{equation}
\begin{split}
    L_{adv} (X, Y_N) & = \sum_{\substack k=1}^{\substack K} p_k c_k \log(c_k),
\end{split}
\label{calibrated_KL_div}
\end{equation}
where $p_k$ denotes the nuisance class distribution in the given training set. To our best knowledge, there has been no similar discussion addressing the class imbalance issue in (adversarial) domain adaption among existing ReID works.

\begin{figure}
\begin{center}
    \includegraphics[width=0.9\linewidth]{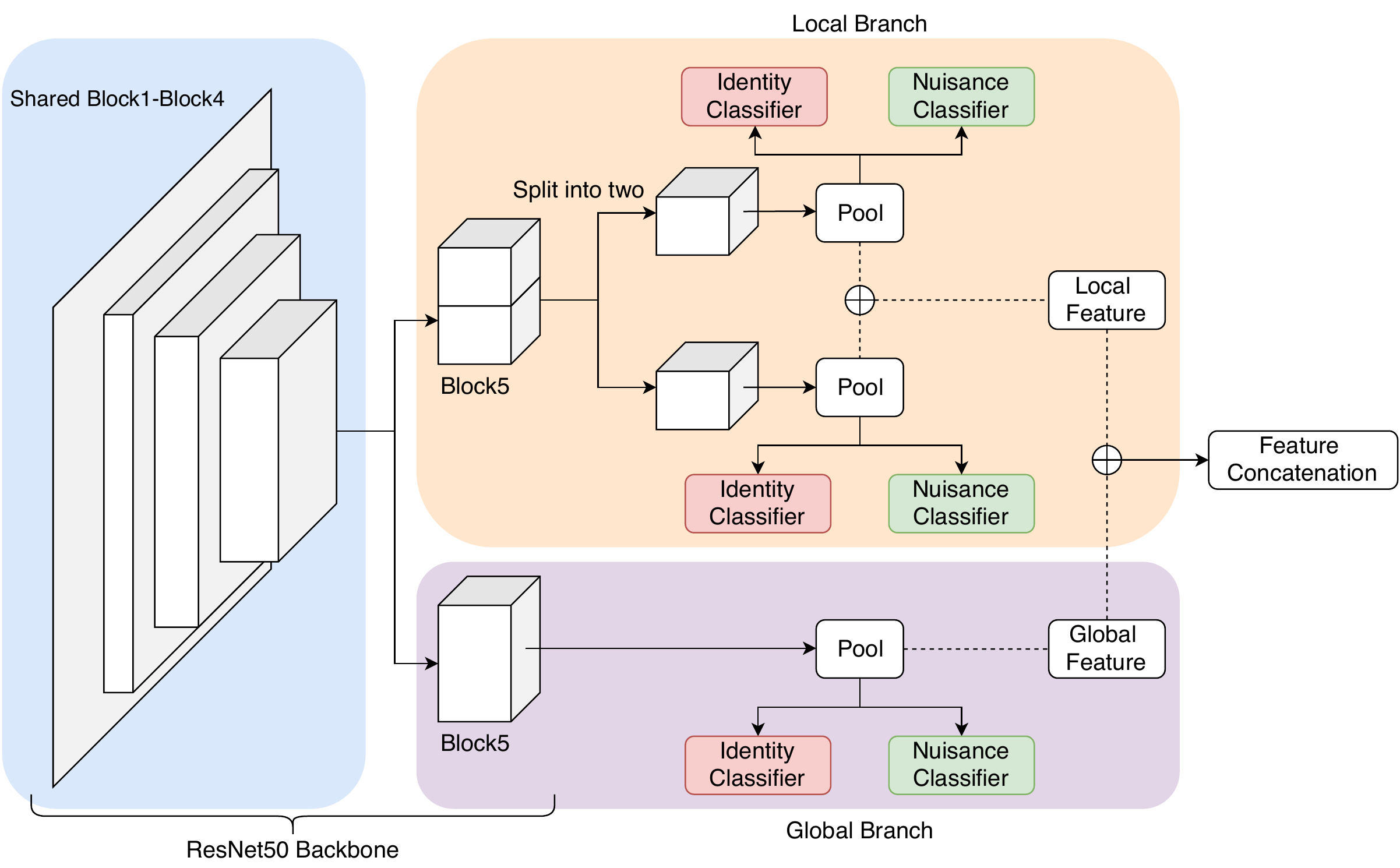}
\end{center}
\vspace{-1em}
\caption{Overview of the dual-branch backbone.}
\vspace{-1em}
\label{fig:dual_branch}
\end{figure}

\begin{table}
\centering
\captionsetup{font=small,justification=centering}
\caption{Ablation study of $L_{adv}$ (direct transfer DukeMTMC-ReID $\shortrightarrow$ Market1501).}
{\scriptsize
\ra{1.0}
\begin{tabular}{cccccc}
\toprule
\multicolumn{2}{c}{\multirow{2}{*}{Settings}} & \multicolumn{4}{c}{DukeMTMC-ReID $\rightarrow$ Market1501} \\ \cline{3-6} 
\multicolumn{2}{c}{} & top1 & top5 & top10 & mAP \\ \midrule
\multicolumn{2}{c}{ResNet50 (baseline)} & 46.8 & 63.5 & 70.3 & 19.0 \\ \midrule
\multirow{3}{*}{\rotatebox[origin=c]{90}{ADIN}} & ResNet50 + Reverse Gradient & \multicolumn{4}{c}{Unable to converge}  \\
 & ResNet50 + NE & 48.8 & 66.2 & 72.7 & 20.4 \\
 & ResNet50 + CaNE & 51.7 & 68.6 & 76.0 & 22.1 \\
 \midrule \midrule
\multicolumn{2}{c}{Dual-branch (baseline)} & 54.8 & 71.7 & 77.6 & 25.9 \\ \midrule
\multirow{3}{*}{\rotatebox[origin=c]{90}{ADIN}} & Dual-branch + Reverse Gradient & \multicolumn{4}{c}{Unable to converge}  \\
 & Dual-branch + NE & 55.9 & {72.5} & {78.6} & 26.5 \\
 & \textbf{Dual-branch + CaNE} & \textbf{57.2} & \textbf{73.0} & \textbf{80.0} & \textbf{27.4} \\
\bottomrule
\end{tabular}
}\label{table:ablation}
\end{table}

\begin{table*}[!tb]
\centering \arraybackslash
\captionsetup{font=normalsize}
\caption{Direct transfer performance from MSMT17 \cite{Wei_2018_CVPR} to DukeMTMC-ReID \cite{Ristani_2016_ECCV, Zheng_2017_ICCV} and to Market1501 \cite{Zheng_2015_ICCV}. \textbf{*} indicates method using images from both source and target domain.
}
{\normalsize
\ra{1.0}
\begin{tabular}{cccccccc}
\toprule
\multirow{2}{*}{} & \multicolumn{3}{c}{MSMT17 $\shortrightarrow$ DukeMTMC-ReID} & \phantom{} & \multicolumn{3}{c}{MSMT17 $\shortrightarrow$ Market1501} \\ \cmidrule{2-4} \cmidrule{6-8}
& top1 & top5 & mAP && top1 & top5 & mAP \\
\midrule
Spatial-Attention\cite{wang2018parameter}  & 52.2 & 68.1 & 32.9 && 49.7 & 68.9 & 25.1 \\
PCB\cite{Sun_2018_ECCV} & 54.4 & 69.6 & 34.6 && 52.7 & 71.3 & 26.7 \\
RPP\cite{Sun_2018_ECCV} & 56.7 & 71.4  & 36.7 && 50.2 & 70.7 & 26.3 \\
MGN\cite{wang2018learning} & 55.5 & 70.2 & 35.1 && 48.7 & 66.9 & 25.1 \\ 
\midrule
HHL\cite{Zhong_2018_ECCV}* & \textbf{62.2} & \textbf{78.8} & 31.4 && 46.9 & 61.0 & 27.2 \\
\midrule
ResNet50 (baseline) & 49.7 & 65.7 & 28.2 && 47.7 & 64.3 & 21.2 \\
ResNet50 + CaNE & 52.6 & 67.9 & 30.4 && 50.1 & 66.4 & 22.5\\ 
\midrule
Dual-branch & 59.5 & 73.5 & 38.4 && 57.8 & 73.9 & 29.4 \\
\textbf{Dual-branch + CaNE} & 60.7 & 74.7 & \textbf{39.1} && \textbf{59.1} & \textbf{75.4} & \textbf{30.3} \\
\bottomrule
\end{tabular} \label{table:direct_transfer_msmt2duke.market}
}
\end{table*}

\subsection{Training Strategy Overview} \label{sec:training}

Figure \ref{fig:model} overviews the concrete training workflow of ADIN, which consists of three modules: feature extractor $f_E$, subject identity classifier $f_I$, and nuisance classifier $f_N$. $f_E$ takes the image $X$ as input and outputs the feature $f_E(X)$, which is then passed through $f_I$ and $f_N$. Both $f_I$ and $f_N$ aim to accurately predict their corresponding labels from the learned features. The training of $f_E$ strives to boost the prediction of $f_I(f_E(X))$, while suppressing the prediction of $f_N(f_E(X))$. It is important to keep $f_N$ strong to maintain a meaningful competition for learning nontrivial $f_E$.

In practice, we implement the training using an iterative strategy.
We initialize $f_E$, $f_I$ and $f_N$ by jointly training the feature extractor $f_E$ and identity classifier $f_I$, and then fixing $f_E$ and pre-training $f_N$ solely on top of that. Afterwards, we alternate between optimizing two sub-problems:
\begin{equation}
\begin{split}
\min_{\substack f_E, f_N} L_{adv}(f_N(f_E(X))),\text{\space\space\space} \min_{\substack f_E, f_I} L_I(f_I (f_E(X)),Y_I).
\end{split}
\label{object_3}
\end{equation}
In each alternating round, we optimize the first objective until the validation error of identity classification reducing below a pre-set $\textit{threshold}_{\text{I-target}}$. We then switch to optimizing the second objective, meanwhile monitoring the resulting changes on the identity classification validation error (since $f_E$ is altered): if it drops below another pre-set $\textit{threshold}_{I-trigger}$, we will switch back to the first object and start the next round of alternations. See supplementary for full training details.

\section{Experiment}
\subsection{Direct Transfer across Datasets as ReID Evaluation Metric} \label{sec:direct_transfer}
In view of current ReID dataset limitations, we propose to choose a new evaluation metric to reflect our ultimate goal of generalizability. 
In real large-scale ReID applications, one would expect a trained model to ``automatically'' scale up to as many unseen, different scenarios as possible. Since different existing ReID datasets were collected in very diverse settings, we propose to use the \textbf{direct transfer} performance, by directly applying a ReID model trained on one dataset's training set onto another's testing set to measure accuracy/mAP, \textbf{without any re-training or adaptation}. The direct transfer performance explicitly takes into account ``zero-shot'' generalizability and penalizes the overfitting of setting/scene-specific nuisances. We argue that it should be as important a performance indicator for ReID as traditional accuracy/mAP on the same dataset, if not more. 


\subsection{Datasets and Model Implementation}

We first perform ``small-scale'', proof-of-concept ablation studies using DukeMTMC-ReID \cite{Ristani_2016_ECCV, Zheng_2017_ICCV} and Market-1501 \cite{Zheng_2015_ICCV} (section \ref{table:ablation}), then go to two large-scale datasets: MSMT17 \cite{Wei_2018_CVPR} for person ReID, and VeRi-776 \cite{liu2016deep} for vehicle ReID (section \ref{sec:direct_transfer}). For all datasets, we use target subject IDs (person or vehicle) as $Y_I$. 
For $Y_N$, we take: (i) camera IDs on DukeMTMC-ReID and on VeRi-776; and (ii) both camera IDs and video timestamps on MSMT17, through a multi-label adversarial classifier.


As a general framework, ADIN can take any backbone for $f_E$, $f_I$ and $f_N$. In section \ref{sec:ablation}, we first test our ADIN with $f_E$ being a basic ResNet50 \cite{he2016deep} to illustrate the effectiveness of our adversarial training. Afterwards, we adopt a more sophisticated dual-branch feature extractor for $f_E$, as inspired by \cite{Sun_2018_ECCV,wang2018learning, chen2019abd}, to demonstrate further boosted performance over state-of-the-arts. The configuration of the dual-branch model is depicted in Fig.\ref{fig:dual_branch}. On top of the $f_E$, we append two simple classifiers as $f_I$ and $f_N$ (Fig. \ref{fig:model}), either taking two fully connected layers. $L_I$ is always implemented using the hybrid loss of cross-entropy and center loss \cite{Wen_2016_ECCV}. An ablation study of $L_{adv}$ is presented in section \ref{sec:loss}; after that, the Calibrated Negative Entropy (CaNE) loss will be our default $L_adv$ unless otherwise specified. 

More details of our models can be found in supplementary. All codes and pre-trained models will be released upon acceptance.

\begin{table*}[!tb]
\centering \arraybackslash
\captionsetup{font=normalsize}
\caption[Caption for LOF]{Direct transfer performance from VeRi-776 \cite{liu2016large} to VehicleID\cite{liu2016deep2}. \textbf{*} indicates method using images from both source and target domain.
}
{\normalsize
\ra{1.0}
\begin{tabular}{cccccccc}
\toprule
\multirow{2}{*}{Method} & \multicolumn{3}{c}{Test size = 1600} & \phantom{} & \multicolumn{3}{c}{Test size = 3200} \\ \cmidrule{2-4} \cmidrule{6-8}
 &  top1 & top5 & mAP &&  top1 & top5 & mAP \\ \midrule
RAM\cite{liu2018ram} & 30.5 & 49.5 & 39.5 && 24.5 & 40.3 & 32.4 \\ 
Spatial-Attention\cite{wang2018parameter} & 39.5 & 57.2 & 47.9  &&  33.7 & 49.6 & 41.6   \\
PCB\cite{Sun_2018_ECCV} & 41.3 & 58.8 & 49.7  &&  35.4 & 51.4 & 43.2  \\
RPP\cite{Sun_2018_ECCV} & 40.6 & 58.4 & 49.1  && 35.0 & 51.1 & 42.9 \\
MGN\cite{wang2018learning} & 39.9 & 62.4 & 50.6 && 32.7 & 53.1 & 42.7 \\\midrule
DAVR\cite{peng2019cross}* & 45.2 & 64.0 & 49.7  && 38.7 & 55.9 & 42.9 \\ 
\midrule
ResNet50 (baseline) & 42.3 & 58.5 & 46.2 && 36.1 & 52.2 & 39.9 \\
ResNet50 + CaNE  & 43.3 & 59.7 & 47.2 && 37.0 & 53.4 & 40.9 \\
\midrule
Dual-branch & 47.3 & 65.3 & 51.6 && 41.2 & 57.9 & 45.3  \\
\textbf{Dual-branch + CaNE} & \textbf{48.7}  & \textbf{67.3} & \textbf{53.1} && \textbf{42.1} & \textbf{59.5} & \textbf{46.3} \\ \bottomrule
\end{tabular} \label{table:direct_transfer_veri}
}
\end{table*}

\subsection{Ablation study of the adversarial loss $L_{adv}$} \label{sec:ablation}

Table \ref{table:ablation} displays a step-by-step comparison for choosing $L_{adv}$, with the direct transfer performance from DukeMTMC-ReID (source domain) to Market1501 (target domain) as the indicator.
Without the adversarial domain-invariant training, both the ResNet50 and Dual-branch backbones achieve low direct transfer accuracy, due to the domain discrepancy across two datasets. We also empirically observe the adversarial effect provided by the reverse gradient (RG) hard to converge, owing to the gradient vanishing/explosion and its sensitivity to the loss magnitude from the nuisance classifier $f_N$. With the negative entropy (NE) loss, our adversarial domain-invariant training forces the entropy of the nuisance classifier's prediction to be maximized, leading to nuisance-uninformative features learned by the feature extractor $f_E$ and reliable direct transfer performance. More importantly, as pointed out in section \ref{sec:intro}, the sampling of nuisances are imbalanced, which intrinsically results in imbalanced levels of adversarial effects on each nuisance. Thus our proposed calibrated negative loss (CaNE) further enables the adversarial training to be attentive w.r.t. different nuisances frequencies. Table \ref{table:ablation} shows that both backbones benefit most from our proposed CaNE adversarial loss.

It is worth noting that even trained within a small-scale domain like DukeMTMC-ReID, the generalizability of both of the two backbones can be boosted by ADIN. 


\begin{figure*}[htb!]
\begin{center}
    \includegraphics[width=.85\linewidth]{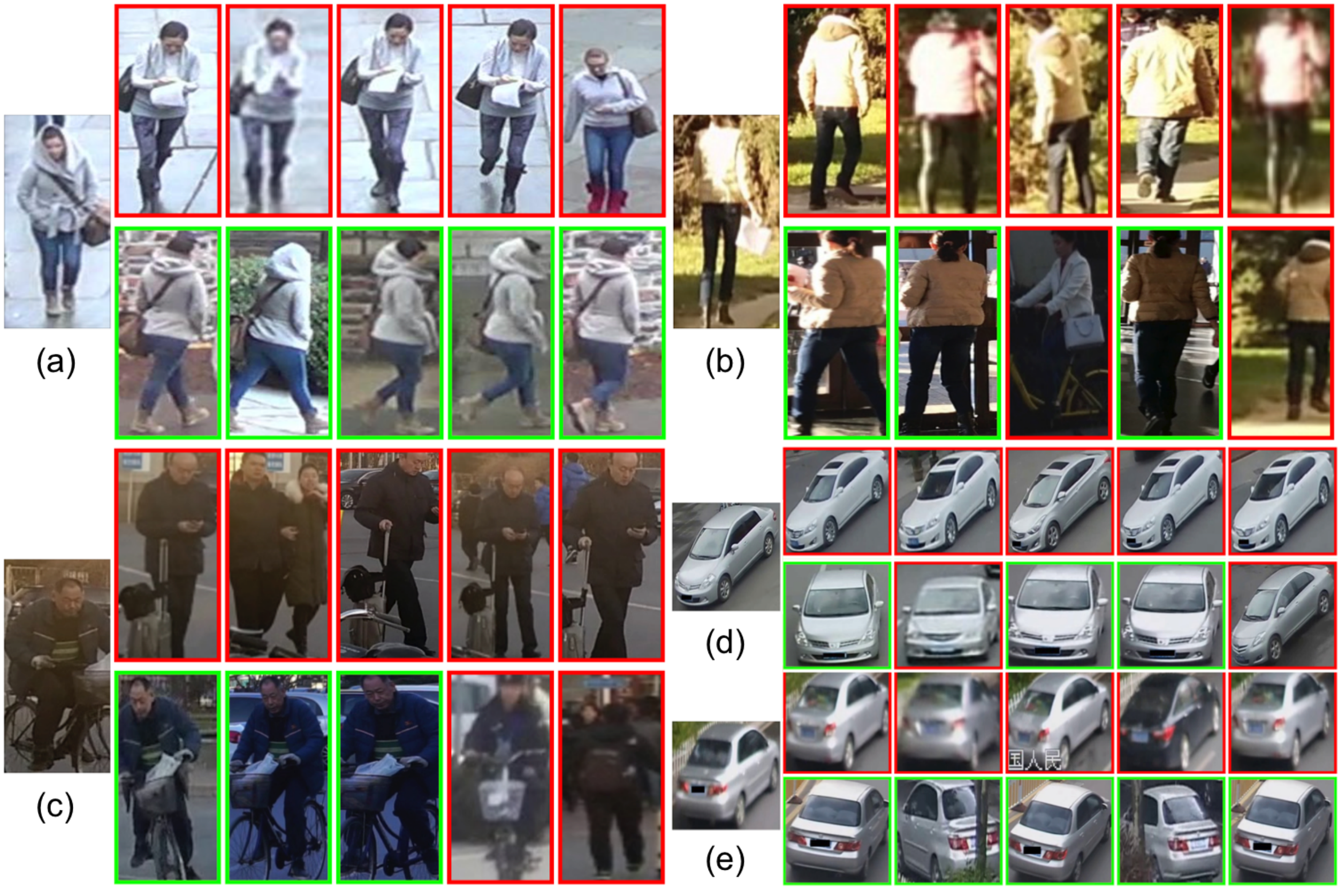}
\end{center}
\vspace{-1em}
\caption{Retrieval results on the DukeMTMC-ReID (a), MSMT17 (b, c), and VeRi-776 (d, e). The leftmost image in each panel is a query image. For the five columns in each panel, the top row shows top-5 retrieval results using a vanilla ResNet50 model \cite{he2016deep}, and the bottom row shows top-5 results using ResNet50 adopted with ADIN framework. Green boxes mark the correct matches, while red boxes denote the wrong matches. The vanilla ResNet50 tend to retrieve images with similar nuisances (illumination, viewpoints, scene backgrounds, etc.), while ADIN successfully eliminates nuisances and leads to more robust matching under drastic visual appearance changes.}
\vspace{-1em}
\label{fig:showcases}
\end{figure*}

\subsection{Direct Transfer between Datasets without Retraining or Adaption}

We evaluate three direct transfer cases, two on person ReID: MSMT17 $\rightarrow$ DukeMTMC-ReID, MSMT-17 $\rightarrow$ Market1501; and one on vehicle ReID: VeRi-776 \cite{liu2016large} $\rightarrow$ VehicleID \cite{liu2016deep2}. As comparison baselines, we train the same dual-branch backbones (without any adversarial learning) on the source datasets, and test their direct transfer performance too. We train and compare with several state-of-the-art ReID models on MSMT17: Spatial-Attention \cite{wang2018parameter}, PCB \cite{Sun_2018_ECCV}, RPP \cite{Sun_2018_ECCV}, MGN \cite{wang2018learning} (Person ReID); and RAM \cite{liu2018ram} (Vehicle ReID). We further compare with
existing best performers of domain adaptation: HHL \cite{Zhong_2018_ECCV} (Person ReID), and DAVR \cite{peng2019cross} (Vehicle ReID), which reported the current best transfer results between DukeMTMC-ReID/Market1501, and from VeRi-776 \cite{liu2016large} to VehicleID \cite{liu2016deep2}, respectively. Note that both HHL and DAVR need to use (unlabeled) target domain data and perform extra (re-)training for the source domain models, while ours need not: the comparisons are thus apparently \underline{to our competitors' advantage}.

As can be seen from Tables \ref{table:direct_transfer_msmt2duke.market} and \ref{table:direct_transfer_veri}, while baselines without adversarial learning fail to transfer well as expected, ADIN demonstrates highly impressive results on all three transfer cases. In particular, by training on MSMT17 and directly transferring, ADIN not only surpasses the direct transfer results from other methods but also outperforms state-of-the-art ReID 
domain adaption model HHL \cite{Zhong_2018_ECCV} and DAVR \cite{peng2019cross}, while costing literally none of their hassles such as (re-)training.

In contrast to our ADIN, we find other (single-dataset) top-performers \cite{wang2018parameter,Sun_2018_ECCV,wang2018learning} generalize very poorly to unseen domains, indicating the misaligned goals between overfitting small-scale single dataset, and generalizing to large-scale unseen scenarios in real life. As in Fig.\ref{fig:scatter_plot}, our ADIN framework lies in the top-right corner, while others stay in the left region with high single-dataset accuracy but poor direct transfer performance. We believe the effective direct transfer is the right choice for evaluating and promoting larger-scale ReID practice, and hope our proposals and arguments could invoke more discussions in the community. We include the full detailed results in the supplementary.

Figure \ref{fig:showcases} shows five visual retrieval examples. In both queries, the spatiotemporal nuisances (\textit{e.g.} the door of same geo-location, certain lighting condition or glare) have a strong presence in images. As can be seen in the top row of each case, the baselines overfit background, tending to retrieve images with similar nuisances (illumination, viewpoints, scene backgrounds, etc.). In contrast, ADIN successfully eliminates them, and leads to much more robust matching under drastic visual appearance changes, as seen in the bottom rows.

In sum, ADIN proves to make a substantial improvement in overcoming the generalizability challenge: the ADIN feature extractor learned on one dataset is directly generalizable to others, without seeing or adapting on any new data. To our best knowledge, ADIN is the first CNN-based ReID model that can establish strong direct transfer performance. It sets up state-of-the-art generalizability for ReID, which we believe is valuable for pursuing real-world large-scale ReID applications.


\section{Conclusion}
This paper proposes the adversarial domain-invariant (\textbf{ADIN}) learning framework, which remarkably improves the generalizability of ReID models and resolves the nuisances-overfitting problem. Free annotations like video timestamp and camera index are for the first time utilized. In extensive experiments, measured by the new \textbf{direct transfer} performance criterion, ADIN exhibits impressive generalization to unseen datasets without any fine-tuning or re-trainig. We hope to draw more research interests to address the ReID generalizability and to pursue more effective direct transfer.

{\small
\bibliographystyle{unsrt}
\bibliography{egbib}

\begin{thebibliography}{10}

\bibitem{Wei_2018_CVPR}
Longhui Wei, Shiliang Zhang, Wen Gao, and Qi~Tian.
\newblock Person transfer gan to bridge domain gap for person
  re-identification.
\newblock In {\em The IEEE Conference on Computer Vision and Pattern
  Recognition (CVPR)}, June 2018.

\bibitem{camera2014}
Number of video surveillance cameras per thousand people in 2014, by country.
\newblock {\em
  https://www.statista.com/statistics/484956/number-of-surveillance-cameras-per-thousand-people-by-country/},
  2015.

\bibitem{wang2018person}
Yicheng Wang, Zhenzhong Chen, Feng Wu, and Gang Wang.
\newblock Person re-identification with cascaded pairwise convolutions.
\newblock In {\em Proceedings of the IEEE Conference on Computer Vision and
  Pattern Recognition}, pages 1470--1478, 2018.

\bibitem{wu2018towards}
Zhenyu Wu, Zhangyang Wang, Zhaowen Wang, and Hailin Jin.
\newblock Towards privacy-preserving visual recognition via adversarial
  training: A pilot study.
\newblock In {\em Proceedings of the European Conference on Computer Vision
  (ECCV)}, pages 606--624, 2018.

\bibitem{wang2019privacy}
Haotao Wang, Zhenyu Wu, Zhangyang Wang, Zhaowen Wang, and Hailin Jin.
\newblock Privacy-preserving deep visual recognition: An adversarial learning
  framework and a new dataset.
\newblock {\em arXiv preprint arXiv:1906.05675}, 2019.

\bibitem{Zheng_2015_ICCV}
Liang Zheng, Liyue Shen, Lu~Tian, Shengjin Wang, Jingdong Wang, and Qi~Tian.
\newblock Scalable person re-identification: A benchmark.
\newblock In {\em The IEEE International Conference on Computer Vision (ICCV)},
  December 2015.

\bibitem{Ristani_2016_ECCV}
Ergys Ristani, Francesco Solera, Roger Zou, Rita Cucchiara, and Carlo Tomasi.
\newblock Performance measures and a data set for multi-target, multi-camera
  tracking.
\newblock In {\em The European Conference on Computer Vision (ECCV)}, September
  2016.

\bibitem{Zheng_2017_ICCV}
Zhedong Zheng, Liang Zheng, and Yi~Yang.
\newblock Unlabeled samples generated by gan improve the person
  re-identification baseline in vitro.
\newblock In {\em The IEEE International Conference on Computer Vision (ICCV)},
  Oct 2017.

\bibitem{wang2018parameter}
Haoran Wang, Yue Fan, Zexin Wang, Licheng Jiao, and Bernt Schiele.
\newblock Parameter-free spatial attention network for person
  re-identification.
\newblock {\em arXiv preprint arXiv:1811.12150}, 2018.

\bibitem{Sun_2018_ECCV}
Yifan Sun, Liang Zheng, Yi~Yang, Qi~Tian, and Shengjin Wang.
\newblock Beyond part models: Person retrieval with refined part pooling (and a
  strong convolutional baseline).
\newblock In {\em The European Conference on Computer Vision (ECCV)}, September
  2018.

\bibitem{wang2018learning}
Guanshuo Wang, Yufeng Yuan, Xiong Chen, Jiwei Li, and Xi~Zhou.
\newblock Learning discriminative features with multiple granularities for
  person re-identification.
\newblock In {\em 2018 ACM Multimedia Conference on Multimedia Conference},
  pages 274--282. ACM, 2018.

\bibitem{Zhong_2018_ECCV}
Zhun Zhong, Liang Zheng, Shaozi Li, and Yi~Yang.
\newblock Generalizing a person retrieval model hetero- and homogeneously.
\newblock In {\em The European Conference on Computer Vision (ECCV)}, September
  2018.

\bibitem{li2014deepreid}
Wei Li, Rui Zhao, Tong Xiao, and Xiaogang Wang.
\newblock Deepreid: Deep filter pairing neural network for person
  re-identification.
\newblock In {\em CVPR}, 2014.

\bibitem{liu2016large}
Xinchen Liu, Wu~Liu, Huadong Ma, and Huiyuan Fu.
\newblock Large-scale vehicle re-identification in urban surveillance videos.
\newblock In {\em 2016 IEEE International Conference on Multimedia and Expo
  (ICME)}, pages 1--6. IEEE, 2016.

\bibitem{liu2016deep2}
Hongye Liu, Yonghong Tian, Yaowei Yang, Lu~Pang, and Tiejun Huang.
\newblock Deep relative distance learning: Tell the difference between similar
  vehicles.
\newblock In {\em Proceedings of the IEEE Conference on Computer Vision and
  Pattern Recognition}, pages 2167--2175, 2016.

\bibitem{liu2016deep}
Xinchen Liu, Wu~Liu, Tao Mei, and Huadong Ma.
\newblock A deep learning-based approach to progressive vehicle
  re-identification for urban surveillance.
\newblock In {\em European Conference on Computer Vision}, pages 869--884.
  Springer, 2016.

\bibitem{camera_amazon}
Amazon go debuts, and its prying cameras foil our shoplifting attempts.
\newblock {\em
  https://arstechnica.com/information-technology/2018/01/we-test-the-worlds-first-amazon-go-watch-you-shop-grocery-store/},
  2018.

\bibitem{visitor_amazon}
Analysts: Amazon go stores bring in 50\% more revenue than typical c-stores.
\newblock {\em
  https://csnews.com/analysts-amazon-go-stores-bring-50-more-revenue-typical-c-stores},
  2019.

\bibitem{Wang_2018_CVPR}
Jingya Wang, Xiatian Zhu, Shaogang Gong, and Wei Li.
\newblock Transferable joint attribute-identity deep learning for unsupervised
  person re-identification.
\newblock In {\em The IEEE Conference on Computer Vision and Pattern
  Recognition (CVPR)}, June 2018.

\bibitem{Tian_2018_CVPR}
Maoqing Tian, Shuai Yi, Hongsheng Li, Shihua Li, Xuesen Zhang, Jianping Shi,
  Junjie Yan, and Xiaogang Wang.
\newblock Eliminating background-bias for robust person re-identification.
\newblock In {\em The IEEE Conference on Computer Vision and Pattern
  Recognition (CVPR)}, June 2018.

\bibitem{Zhong_2018_CVPR}
Zhun Zhong, Liang Zheng, Zhedong Zheng, Shaozi Li, and Yi~Yang.
\newblock Camera style adaptation for person re-identification.
\newblock In {\em The IEEE Conference on Computer Vision and Pattern
  Recognition (CVPR)}, June 2018.

\bibitem{Ma_2018_CVPR}
Liqian Ma, Qianru Sun, Stamatios Georgoulis, Luc Van~Gool, Bernt Schiele, and
  Mario Fritz.
\newblock Disentangled person image generation.
\newblock In {\em The IEEE Conference on Computer Vision and Pattern
  Recognition (CVPR)}, June 2018.

\bibitem{Deng_2018_CVPR}
Weijian Deng, Liang Zheng, Qixiang Ye, Guoliang Kang, Yi~Yang, and Jianbin
  Jiao.
\newblock Image-image domain adaptation with preserved self-similarity and
  domain-dissimilarity for person re-identification.
\newblock In {\em The IEEE Conference on Computer Vision and Pattern
  Recognition (CVPR)}, June 2018.

\bibitem{peng2019cross}
Jinjia Peng, Huibing Wang, and Xianping Fu.
\newblock Cross domain knowledge learning with dual-branch adversarial network
  for vehicle re-identification.
\newblock {\em arXiv preprint arXiv:1905.00006}, 2019.

\bibitem{ganin2014unsupervised}
Yaroslav Ganin and Victor Lempitsky.
\newblock Unsupervised domain adaptation by backpropagation.
\newblock {\em arXiv preprint arXiv:1409.7495}, 2014.

\bibitem{Yaroslav_2016_JMLR}
Yaroslav Ganin, Evgeniya Ustinova, Hana Ajakan, Pascal Germain, Hugo
  Larochelle, Fran\c{c}ois Laviolette, Mario Marchand, and Victor Lempitsky.
\newblock Domain-adversarial training of neural networks.
\newblock {\em Journal of Machine Learning Research}, 17(59):1--35, 2016.

\bibitem{lin2017focal}
Tsung-Yi Lin, Priya Goyal, Ross Girshick, Kaiming He, and Piotr Doll{\'a}r.
\newblock Focal loss for dense object detection.
\newblock In {\em Proceedings of the IEEE international conference on computer
  vision}, pages 2980--2988, 2017.

\bibitem{he2016deep}
Kaiming He, Xiangyu Zhang, Shaoqing Ren, and Jian Sun.
\newblock Deep residual learning for image recognition.
\newblock In {\em Proceedings of the IEEE conference on computer vision and
  pattern recognition}, pages 770--778, 2016.

\bibitem{chen2019abd}
Tianlong Chen, Shaojin Ding, Jingyi Xie, Ye~Yuan, Wuyang Chen, Yang Yang, Zhou
  Ren, and Zhangyang Wang.
\newblock Abd-net: Attentive but diverse person re-identification.
\newblock {\em ICCV}, 2019.

\bibitem{Wen_2016_ECCV}
Yandong Wen, Kaipeng Zhang, Zhifeng Li, and Yu~Qiao.
\newblock A discriminative feature learning approach for deep face recognition.
\newblock In {\em The European Conference on Computer Vision (ECCV)}, pages
  499--515. Springer, 2016.

\bibitem{liu2018ram}
Xiaobin Liu, Shiliang Zhang, Qingming Huang, and Wen Gao.
\newblock Ram: a region-aware deep model for vehicle re-identification.
\newblock In {\em 2018 IEEE International Conference on Multimedia and Expo
  (ICME)}, pages 1--6. IEEE, 2018.

\end{thebibliography}
}

\end{document}


\title{Supplementary Materials: Calibrated Domain-Invariant Learning for\\Highly Generalizable Large Scale Re-Identification}

\author{First Author \\
Institution1\\
{\tt\small firstauthor@i1.org}
\and
Second Author \\
Institution2\\
{\tt\small secondauthor@i2.org}
}

\maketitle
\ifwacvfinal\thispagestyle{empty}\fi

\section{Training Details}


To start our adversarial domain-invariant learning (ADIN), we initialize $f_E$, $f_I$ and $f_N$ with pretrained weights. We first jointly pretrain the feature extractor $f_E$ and identity classifier $f_I$, and pretrain $f_N$ by fixing $f_E$. In ADIN we alternative between optimizing two sub-problems:
\begin{equation}
\begin{split}
\min_{\substack f_E, f_N} L_{adv}(f_N(f_E(X))), \min_{\substack f_E, f_I} L_I(f_I (f_E(X)),Y_I).
\end{split}
\label{object_3}
\end{equation}
In each alternating round, we optimize the first objective until the validation error of identity classification reducing below a pre-set $threshold_{I-target}$. We then switch to optimizing the second objective, meanwhile monitoring the resulting changes on the identity classification validation error (since $f_E$ is updated): if it drops below another pre-set $threshold_{I-trigger}$, we will switch back to the first object and start the next round of alternations. 

Besides, to meet the ``hidden constraint'' and avoid too weak $f_N$ during training, we will periodically replace the current weights in $f_I$ and $f_N$ with random weights, and re-train them on $f_E(X)$ by:
\begin{equation}
\begin{split}
\min_{\substack f_N} L_N(f_N (f_E(X)),Y_I), \min_{\substack f_I} L_I(f_I (f_E(X)),Y_I).
\end{split}
\label{object_4}
\end{equation}
with $f_E$ being unaffected and fixed. We then re-start alternations from the new ``pre-trained'' initialization. This empirical trick is found to strengthen the generalization of $f_E$, potentially because it gets rid of some trivial local minima. The training procedure is summarized in Algorithm \ref{training}.


\begin{algorithm*}[h!]
\caption{The training strategy of ADIN}\label{adv_training}
\begin{algorithmic}[1]
{\small
\State Given pre-trained feature extractor $f_E$, identity classifier $f_I$  and nuisances classifier $f_N$
\State $val_I, val_N \gets$ {identity classifier validation accuracy},  {nuisances classifier validation accuracy}.
\For {number of training epoches}
    \If{$val_I < threshold_{I-trigger}$} 
    \Comment{Avoid weak identity recognition performance}
        \While {$val_I \leq threshold_{I-target}$}
            \For {number of batches}
                \State Sample minibatch of m examples $\{ X_1, ... , X_m \}$
                \State Jointly update the $f_E$ and the $f_I$ by descending its stochastic gradient with loss $L_I$
            \EndFor
            \State $val_I \gets$ {identity classifier validation accuracy}.
        \EndWhile
    \ElsIf {$val_N > threshold_{N}$} 
    \Comment{Suppress nuisance discriminator performance}
        \State Feed all training examples $\{ X_1, ... , X_n \}$ into the model
        \State Jointly update $f_E$ and $f_N$ by descending its gradient with the adversarial loss $L_{adv}$
        
    \Else \Comment{Further boost identity recognition performance}
        \For {number of batches}
            \State Sample minibatch of m examples $\{ X_1, ... , X_m \}$
            \State Jointly update $f_E$ and $f_I$ by descending its stochastic gradient with loss $L_I$
        \EndFor
    \EndIf 
    \State Re-initialize $f_I$, $f_N$ 
    \Comment{Empirically restart $f_I$ $f_N$ every iteration to avoid overfitting extracted features}
    \State Train $f_I$, $f_N$ by descending its stochastic gradient with classification loss $L_I$, $f_N$ correspondingly
    \State $val_I, val_N \gets$ {identity classifier validation accuracy}, {nuisances classifier validation accuracy}.
\EndFor
}
\end{algorithmic}
\label{training}
\end{algorithm*}

\section{Network Structure of Dual-branch Backbone}
\begin{figure}
\begin{center}
    \includegraphics[width=0.9\linewidth]{wacv2020-author-kit/latex/figures/dual-branch.pdf}
\end{center}
\caption{Overview of the dual-branch backbone.}
\label{fig:dual_branch}
\end{figure}
For our dual-branch backbone, the first four blocks share the same design as in ResNet50. After the forth block, the network was split into a global and a local branch. In the global branch, the feature passes a global average-pooling and then is fed into the classifier. In the local branch, feature is horizontally partitioned into two equal parts, where each part adopts a separate global average-pooling layer and classifier. During inference the outputs from two branches are concatenated together as the final feature for image retrieval.

\section{Additional Experiment Results}
\subsection{Ablation study of the adversarial loss $L_{adv}$}

\begin{table}[h!]
\centering
\caption{Ablation study of adversarial loss: single-dataset performance on DukeMTMC-ReID \cite{Ristani_2016_ECCV, Zheng_2017_ICCV} and direct transfer of DukeMTMC-ReID $\rightarrow$ Market1501 \cite{Zheng_2015_ICCV}. \textbf{*} indicates method using images from both source and target domain.}
{\scriptsize
\ra{1.0}
\resizebox{\columnwidth}{!}{
\begin{tabular}{cccccccccc}
\toprule

\multirow{2}{*}{} & \multicolumn{4}{c}{DukeMTMC-ReID} & \phantom{} & \multicolumn{4}{c}{DukeMTMC-ReID $\rightarrow$ Market1501} \\ \cmidrule{2-5} \cmidrule{7-10}
& top1 & top5 & top10 & mAP && top1 & top5 & top10 & mAP \\
\midrule
Spatial-Attention\cite{wang2018parameter} & 83.6 & 91.8 & \textbf{94.3} & 70.1 && 49.8 & 67.4 & 73.8 & 24.2 \\
PCB\cite{Sun_2018_ECCV} & \textbf{84.2} & 91.7 & 93.4 & 69.7 && 53.9 & 70.5 & 76.8 & 26.5\\
RPP\cite{Sun_2018_ECCV} & 84.1 & \textbf{92.5} & \textbf{94.3} & \textbf{71.3} && 53.1 & 71.3 & 76.6 & 25.7 \\
MGN\cite{wang2018learning} & 55.5 & 70.2 & 76.8 & 35.1 && 48.7 & 66.9 & 73.7 & 25.1 \\ 
\midrule
CycleGAN \cite{Deng_2018_CVPR} * & - & - & - & - && 48.1 & 66.2 & 72.7 & 20.7 \\
SPGAN \cite{Deng_2018_CVPR} * & - & - & - & - && 58.1 & 76.0 & 82.7 & 26.9 \\
HHL \cite{Zhong_2018_ECCV} * & - & - & - & - && \textbf{62.2} & \textbf{78.8} & \textbf{84.0} & \textbf{31.4} \\
\midrule
ResNet50 (baseline) & 76.6 & 88.0 & 91.5 & 58.1 && 46.8 & 63.5 & 70.3 & 19.0 \\ 
ResNet50 + Reverse Gradient & \multicolumn{9}{c}{Unable to converge}  \\
ResNet50 + NE & 76.7 & 87.9 & 91.2 & 57.5 && 48.8 & 66.2 & 72.7 & 20.4 \\
ResNet50 + CaNE & 74.8 & 86.0 & 89.1 & 54.9 && 51.7 & 68.6 & 76.0 & 22.1  \\ 
\midrule
Dual-branch & 82.1 & 91.3 & 93.0 & 66.8 && 54.8 & 71.7 & 77.6 & 25.9 \\
Dual-branch + Reverse Gradient & \multicolumn{9}{c}{Unable to converge}  \\
Dual-branch + NE & 81.8 & 90.8 & 93.3 & 66.4 && 55.9 & 72.5 & 78.6 & 26.5 \\
Dual-branch + CaNE & 80.7 & 90.0 & 91.9 & 63.8 && 57.2 & 73.0 & 80.0 & 27.4 \\ 
\toprule
\end{tabular}
}}\label{table:ablation}
\end{table}

Table \ref{table:ablation} 
displays a full step-by-step comparison for different adversarial loss $L_{adv}$, together with state-of-the-art ReID models and domain adaptation methods. We use the direct transfer performance from DukeMTMC-ReID (source domain) to Market1501 (target domain) as the indicator.

\begin{table*}[ht!]
\centering \arraybackslash
\caption{Single-dataset performance on MSMT17 \cite{Wei_2018_CVPR} and direct transfer of MSMT17 $\rightarrow$ DukeMTMC-ReID and MSMT17 $\rightarrow$ Market1501. \textbf{*} indicates method using images from both source and target domain.}
{\scriptsize
\ra{1.0}
\begin{tabular}{cp{0.3cm}p{0.3cm}p{0.3cm}p{0.3cm}ccccccp{0.4cm}p{0.4cm}p{0.4cm}p{0.4cm}}
\toprule
\multirow{2}{*}{} & \multicolumn{4}{c}{MSMT17} & \phantom{} & \multicolumn{4}{c}{MSMT17 $\rightarrow$ DukeMTMC-ReID} & \phantom{} & \multicolumn{4}{c}{MSMT17 $\rightarrow$ Market1501} \\ \cmidrule{2-5} \cmidrule{7-10} \cmidrule{12-15} 
& top1 & top5 & top10 & mAP && top1 & top5 & top10 & mAP && top1 & top5 & top10 & mAP \\
\midrule
Spatial-Attention\cite{wang2018parameter} & 68.7 & 81.5 & 85.7 & 41.8 && 52.2 & 68.1 & 74.1 & 32.9 && 49.7 & 68.9 & 75.5 & 25.1 \\
PCB\cite{Sun_2018_ECCV} & 68.6 & 81.3 & 85.8 & 41.8 && 54.4 & 69.6 & 75.4 & 34.6 && 52.7 & 71.3 & 77.5 & 26.7 \\
RPP\cite{Sun_2018_ECCV} & 73.1 & \textbf{84.5} & \textbf{88.1} & \textbf{46.4} && 56.7 & 71.4 & 76.8 & 36.7 && 50.2 & 70.7 & 77.5 & 26.3 \\
MGN\cite{wang2018learning} & 71.7 & 83.3 & 87.1 & 45.7 && 55.5 & 70.2 & 76.8 & 35.1 && 48.7 & 66.9 & 73.7 & 25.1 \\ 
\midrule
CycleGAN \cite{Deng_2018_CVPR} * & - & - & - & - &
 & 48.1 & 66.2 & 72.7 & 20.7 && 38.5 & 54.6 & 60.8 & 19.9 \\
SPGAN \cite{Deng_2018_CVPR} * & - & - & - & - &
 & 58.1 & 76.0 & 82.7 & 26.9 && 46.9 & 62.6 & 68.5 & 26.4\\
HHL \cite{Zhong_2018_ECCV} * & - & - & - & - &
 & \textbf{62.2} & \textbf{78.8} & \textbf{84.0} & 31.4 && 46.9 & 61.0 & 66.7 & 27.2 \\
\midrule
ResNet50 (baseline) & 63.2 & 76.7 & 81.6 & 31.9 && 49.7 & 65.7 & 71.0 & 28.2 && 47.7 & 64.3 & 71.5 & 21.2 \\
ResNet50 + CaNE & 62.7 & 76.5 & 81.0 & 30.8 && 52.6 & 67.9 & 73.2 & 30.4 && 50.1 & 66.4 & 73.5 & 22.5 \\
\midrule
Dual-branch & \textbf{73.5} & 84.3 & \textbf{88.1} & 45.1 && 59.5 & 73.5 & 78.8 & 38.4 && 57.8 & 73.9 & 80.6 & 29.4\\
Dual-branch + CaNE  & 73.3 & \textbf{84.5} & 87.8 & 43.0 && 60.7 & 74.7 & 79.5 & \textbf{39.1} && \textbf{59.1} & \textbf{75.4} & \textbf{81.7} & \textbf{30.3} \\
\toprule
\end{tabular} \label{table:msmt}
}
\end{table*}

Compared with the baseline and Reverse Gradient, our proposed calibrated negative loss (CaNE) contributes to both stable adversarial training (robustness against gradient vanishing/explosion and classifier's loss magnitude), and CaNE is also attentive adversarial effects w.r.t. different nuisances frequencies (attention to sampling imbalance in nuisances), outperforming the negative entropy (NE) loss. Our ADIN framework with CaNE loss not only improve the generalizability of basic backbones like ResNet50, but also boosts more powerful ones like Dual-branch, indicating that ADIN is a general effective adversarial learning framework towards generalizability. Impressively, the Dual-branch backbone equipped with ADIN and CaNE outperforms even domain adaptation methods (HHL \cite{Zhong_2018_ECCV}, SPGAN \cite{Deng_2018_CVPR}), where extra target source images and fine-tuning/retraining are required but not in our case. We observed that the ADIN causes a bit decrease in single-dataset accuracy. This is because models with our adversarial training no longer overfitting current small-scale dataset.

\begin{table}[h!]
\centering
\caption{Single-dataset performance on VeRi-776 \cite{liu2016large}.
}
{\scriptsize
\ra{1.0}
\begin{tabular}{ccccc}
\toprule
\multirow{2}{*}{} & \multicolumn{4}{c}{VeRI-776 $\shortrightarrow$ VeRi-776} \\ \cmidrule{2-5}
 & top1 & top5 & top10 & mAP \\ \midrule
MAA\cite{jiang2018multi} & 88.0 & 94.6 & - & 58.1 \\
QD-DLF\cite{zhu2019vehicle} & 88.5 & 94.5 & - & 61.8 \\
RAM\cite{liu2018ram} & 88.6 & 94.0 & - & 61.5 \\
GAN+LSRO\cite{wu2018joint} & 87.7 & 93.9 & - & 58.2 \\
BS\cite{kumar2019vehicle} & 90.2 & 96.4 & - & 67.6 \\
Spatial-Attention\cite{wang2018parameter} & 93.4 & 96.8 & 98.2 & 70.5 \\
PCB\cite{Sun_2018_ECCV} & 92.4 & 96.7 & 98.3 & 69.4 \\
RPP\cite{Sun_2018_ECCV} & 93.5 & 96.9 & 97.9 & 69.7 \\
MGN\cite{wang2018learning} & 94.9 & 97.0 & 97.5 & 78.7\\
\midrule
ResNet50 (baseline) & 91.1 & 95.5 & 97.3 & 60.2 \\
ResNet50 + CaNE & 90.7 & 95.41 & 96.9 & 59.5 \\
\midrule
Dual-branch (baseline) & \textbf{95.0} & \textbf{97.9} & \textbf{98.9} & \textbf{73.5} \\
Dual-branch + CaNE & 93.1 & 96.5 & 97.9 & 69.4 \\
 
\toprule
\end{tabular} \label{table:veri}
}
\end{table}

\subsection{Single-dataset and Direct Transfer Performance without Retraining or Adaption}

\begin{table*}[hb!]
\centering \arraybackslash
\caption{Direct transfer performance between VeRi-776 \cite{liu2016large} and VehicleID\cite{liu2016deep2}. \textbf{*} indicates method using images from both source and target domain.}
{\tiny
\ra{1.0}
\begin{tabular}{p{2.2cm}p{0.15cm}p{0.15cm}p{0.15cm}p{0.3cm}cp{0.15cm}p{0.15cm}p{0.15cm}p{0.3cm}cp{0.15cm}p{0.15cm}p{0.15cm}p{0.3cm}cp{0.15cm}p{0.15cm}p{0.15cm}p{0.3cm}}
\toprule
\multirow{2}{*}{Method} & \multicolumn{4}{c}{Test size = 800} & \phantom{} & \multicolumn{4}{c}{Test size = 1600} & \phantom{} &  \multicolumn{4}{c}{Test size = 2400} & \phantom{} & \multicolumn{4}{c}{Test size = 3200} \\ \cmidrule{2-5} \cmidrule{7-10} \cmidrule{12-15} \cmidrule{17-20}
 & top1 & top5 & top10 & mAP && top1 & top5 & top10 & mAP && top1 & top5 & top10 & mAP && top1 & top5 & top10 & mAP \\ \midrule
RAM\cite{liu2018ram} & 34.0 & 53.7 & 61.9 & 43.3 && 30.5 & 49.5 & 56.4 & 39.5 && 26.6 & 43.1 & 51.1 & 34.8 && 24.5 & 40.3 & 48.2 & 32.4 \\
Spatial-Attention\cite{wang2018parameter} & 42.4 & 61.1 & 69.3 & 51.5 && 39.5 & 57.2 & 64.2 & 47.9 && 36.0 & 52.7 & 60.2 & 44.2 && 33.7 & 49.6 & 56.9 & 41.6 \\
PCB\cite{Sun_2018_ECCV} & 43.7 & 63.1 & 70.8 & 53.0 && 41.3 & 58.8 & 65.2 & 49.7 && 37.5 & 53.9 & 61.5 & 45.6 && 35.4 & 51.4 & 58.3 & 43.2 \\
RPP\cite{Sun_2018_ECCV} & 44.5 & 63.1 & 70.1 & 53.2 && 40.6 & 58.4 & 65.2 & 49.1 && 37.0 & 54.1 & 61.8 & 45.3 && 35.0 & 51.1 & 58.4 & 42.9 \\
MGN\cite{wang2018learning} & 44.6 & 70.5 & \textbf{79.7} & 56.5 && 39.9 & 62.4 & 72.3 & 50.6 && 36.2 & 58.1 & 68.0 & 46.6 && 32.7 & 53.1 & 63.0 & 42.7 \\ \midrule
DAVR\cite{peng2019cross}*  & 49.5 & 68.7 & -- & 54.0 && 45.2 & 64.0 & -- & 49.7 &&  40.7 & 59.0 & -- & 45.2 && 38.7 & 55.9 & -- & 42.9 \\ \midrule
ResNet50 (baseline) & 44.7 & 62.5 & 69.0 & 48.9 && 42.3 & 58.5 & 64.5 & 46.2 && 38.1 & 54.8 & 61.7 & 42.1 && 36.1 & 52.2 & 58.9 & 39.9 \\
ResNet50 + CaNE 
 & 46.0 & 63.6 & 69.9 & 50.2 &
 & 43.3 & 59.7 & 65.6 & 47.2 &
 & 38.8 & 56.0 & 63.1 & 42.9 &
 & 37.0 & 53.4 & 60.0 & 40.9 \\
Dual-branch & 51.2 & 70.3 & 77.8 & 55.7 && 47.3 & 65.3 & 72.2 & 51.6 && 44.2 & 62.1 & 69.7 & 48.4 && 41.2 & 57.9 & 64.9 & 45.3 \\
Dual-branch + CaNE
 & \textbf{52.9} & \textbf{72.1} & 79.4 & \textbf{57.4} &
 & \textbf{48.7} & \textbf{67.3} & \textbf{74.0} & \textbf{53.1} &
 & \textbf{45.0} & \textbf{64.0} & \textbf{71.3} & \textbf{49.5} &
 & \textbf{42.1} & \textbf{59.5} & \textbf{66.6} & \textbf{46.3} \\
\bottomrule
\end{tabular} \label{table:direct_transfer_veri}
}
\end{table*}

Here we include the full detailed results. We evaluate three direct transfer cases, two on person ReID: MSMT17 $\rightarrow$ DukeMTMC-ReID, MSMT-17 $\rightarrow$ Market1501; and one on vehicle ReID: VeRi-776 \cite{liu2016large} $\rightarrow$ VehicleID\cite{liu2016deep2}. As comparison baselines, we train the same dual-branch backbones (without any adversarial learning) on the source datasets, and test their direct transfer performance too. We further compare with 
existing strong competitors: one person ReID, a CycleGAN baseline as adopted by \cite{Deng_2018_CVPR} for learning an unsupervised data-level domain mapping, two state-of-the-art domain adaptation methods SPGAN \cite{Deng_2018_CVPR} and HHL \cite{Zhong_2018_ECCV}, the latter reporting the current best transfer results between DukeMTMC-ReID and Market1501; on vehicle ReID the DAVR \cite{peng2019cross} which reported the current best transfer results from VeRi-776 \cite{liu2016large} to VehicleID \cite{liu2016deep2}. Note that CycleGAN, SPGAN and HHL all need to use (unlabeled) target domain data and perform extra (re-)training for the source domain models: the comparisons are thus apparently \underline{to our competitors' advantage}.

As can be seen from Tables \ref{table:msmt} and \ref{table:direct_transfer_veri}, while baselines without adversarial learning fail to transfer well as expected, ADIN demonstrates highly impressive results on all three transfer cases. In particular, by training on MSMT17 and directly transferring, ADIN not only surpasses the direct transfer results from other methods but also outperforms state-of-the-art ReID 
domain adaption models (HHL \cite{Zhong_2018_ECCV}, SPGAN\cite{Deng_2018_CVPR}, DAVR \cite{peng2019cross})
, while costing literally no hassle such as (re-)training.

However, in contrast to our ADIN, we find other (single-dataset) top-performers generalize very poorly to unseen domains, indicating the misaligned goal between overfitting small-scale single dataset and generalizing to large-scale unseen scenarios in real life. We believe the effective direct transfer is the right choice for evaluating and promoting larger-scale ReID practice, and hope our proposals and arguments could invoke more discussions in the community. Again, we observed that the ADIN causes a bit decrease in single-dataset accuracy, since models with our adversarial training no longer overfitting current small-scale dataset.

{\small
\bibliographystyle{unsrt}
\bibliography{egbib}
}